\documentclass[manuscript,screen,authoryear]{acmart}
\AtBeginDocument{%
  }

\setcopyright{acmlicensed}
\copyrightyear{2027}
\acmYear{2027}
\acmDOI{XXXXXXX.XXXXXXX}
\acmConference[FAccT2027]{Submitted for ACM Fairness, Accountability, Transparency}{2027}{Porto, Portugal}
\usepackage{tabulary}
\usepackage{minted}
\usepackage{xcolor}
\usepackage{tcolorbox}
\usepackage{soul}
\usepackage{multirow}

\begin{document}

\title{From Bias to Positionality: A critical grammar for Large Language Models}


\author{Dave Murray-Rust}
\email{d.s.murray-rust@tudelft.nl}
\author{Elisa Giaccardi}
\author{Federico Bomba}
\author{Maria Luce Lupetti}
%\correspondingauthor
%\email{webmaster@marysville-ohio.com}
%\affiliation{%
  %\institution{Institute for Clarity in Documentation}
  %\country{USA}
%}

\renewcommand{\shortauthors}{Murray-Rust et al.}

\begin{abstract}
Bias has become a default lens for understanding what machine learning systems do and how we should relate to them. Bias is at root a deviation from a standard; in decision making systems we can supply a standard from outside: a law, a protected attribute, a benchmark, a statistical property. However, even there, \textit{biased} is often used as a shorthand, without stating what the position taken is - a silent ``relative to''. In a general use of language models, nothing supplies the standard: there is no decision to audit and no counterfactual output, so the shorthand fails. We argue that positionality — the located, articulated, owned and contestable position from which anything is said — is the better grammar for the question these models now force: who is speaking, from where? Against the vendors' answer of \textit{nowhere} and the seductive possibility of \textit{everywhere}, we argue that when a model is taken \textit{as-if} speaking, it speaks from \textit{somewhere}, and this means positionality can describe models, and the way they are believed beyond their warrant, even though they are not accountable speakers.  A model’s positionality is laminar — laid down across corpus, tuning, system prompt, application and the live conversation. We separate three sets of factors: the positionality of a model’s creators, the behaviour they intend for it, and the behaviour it enacts, and situate current work within this matrix. This lets us highlight what each stratum inherits from the one below, with ruptures and collapses where that inheritance is disavowed. We then return to where models speak from, and discuss the implications for how models are made, read and held accountable.
\end{abstract}


 %\includegraphics[width=\textwidth]{sampleteaser}
  %\caption{Seattle Mariners at Spring Training, 2010.}
  %\Description{Enjoying the baseball game from the third-base
  %seats. Ichiro Suzuki preparing to bat.}
  %\label{fig:teaser}

\received{4 Nov 2026}

\maketitle

\section{Introduction}

Bias is an effective framing for issues with decision making systems: the standard is supplied from outside — by law, a protected attribute, a benchmark, through principles of non-discrimination \cite{gerards2022ProtectedGrounds,zuiderveenborgesius2020StrengtheningLegal}; \cite[Article 4a]{AIomnibus2026}. In AI systems more generally, biases range from this statistical representativeness \cite{zhou2022BiasFairness,buolamwini2018GenderShades,selbst2019FairnessAbstraction,green2020AlgorithmicRealism,raji2020ClosingAI,barocas2019FairnessMachine} through the production of gendered text and images  \cite{guilbeault2024OnlineImages,bolukbasi2016ManComputer,zhao2018GenderBias,bianchi2023EasilyAccessible,prates2020AssessingGender,birhane2024DarkSide,qadri2023AIsRegimes,sheng2019WomanWorked} and into wider cultural productions \cite{kommers2026ComputationalHermeneutics,cremaschi2025ExposingIdeology,noble2018AlgorithmsOppression,bunz2019CalculationMeaning,daryani2025HomogenizingEngine,navneet2026WhenAI,jiang2023AIArt,srinivasan2021BiasesGenerative}. However, bias is often under-defined in the literature \cite{ghosh2025BiasMath}, and it works as shorthand - there is always a silent '... \emph{relative to}'.  As commonly deployed, the term \emph{bias} presupposes \emph{positionality}, but does not clearly articulate it. In this paper, we contend that \emph{bias} is the wrong grammar for what large language models (LLMs) do and how we relate to them. Instead, \emph{positionality} fits better: it is a concept that already comes with practices of owning, articulating, costing and contesting a position, and avoids the presupposition of an \emph{unbiased} position.

In language models, particularly when used for cultural production, there is no clear reference point to build bias relative to. Bias can be measured in limited situations: the insertion of pronouns into template documents \cite{zhao2018GenderBias}, roundtrip translations \cite{prates2020AssessingGender}, choice of gender or style when generating images \cite{bianchi2023EasilyAccessible,guilbeault2024OnlineImages}, ideology \cite{cremaschi2025ExposingIdeology,buyl2024LargeLanguage} – but even discussing ideology brings us into the realm where bias breaks down: there is no \emph{decision} to audit, no \emph{counterfactual output}, no protected attribute to test around. Further, part of cultural production is actually having a voice: the deviation from neutrality is precisely what is valued \cite{kommers2026ComputationalHermeneutics,srinivasan2021BiasesGenerative}. Without a clear standard to build bias relative to, what matters is the position taken in the assemblage of model, system and user.

Language models are increasingly taken \emph{as if} they are speaking – even if they lack the conditions of accountability to be a true \emph{speaker}  \cite{gomez2026AssertionAccountability,zhang2026UnanswerableArguer} – and in this frame it makes sense to ask where the speech comes \emph{from}. The default answer has been \emph{nowhere}: the position of the vendors, that their  models are even-handed, apolitical instruments with no positions of their own, except some \emph{bias} that may be reduced \cite{openai2025DefiningEvaluating,anthropic2025MeasuringPolitical}, as an approximation of \emph{neutrality} \cite{fisher2025PoliticalNeutrality}. In contrast, Neuman \cite{neuman2026JusticeIdentity} argues that language models offer the possibility of a view from \emph{everywhere}, as they contain "representations of many social identities without personally inhabiting any one of them", giving a "moral standpoint unavailable to individual human reasoners". While there is something compelling to the idea that models are "condensates of human culture" \cite{neuman2026JusticeIdentity}, and have a unique point to speak from, we argue it is important to treat it as \emph{somewhere}. LLMs have a  \emph{credibility excess} \cite{fricker2007EpistemicInjustice,bondy2010ArgumentativeInjustice,zhang2026UnanswerableArguer}, where they are seen as more authoritative than warranted; this makes it relevant to ask what is the \emph{somewhere} from which they speak, rather than falling back on the abstracted \emph{nowhere} and \emph{everywhere}.

Positionality keeps the question "who is speaking, from where?" open, framing the shaping of a model as \emph{taking a position} rather than removing a defect. Models take a particular, sedimented, composite position — laid down in a corpus, filtered by raters, written into a specification, steered by a deployer, bent by the conversation — that can be located, articulated, owned and contested, even without "a singular human social position." \cite{neuman2026JusticeIdentity}. We explore what it means to take seriously the idea that a language model has a positionality, and ask what that is.   Philosophically, this draws on a \emph{cosmopolitical} worldview \cite{stengers2005CosmopoliticalProposal}, an avoidance of flattening towards a single consensus through keeping open multiple, divergent worlds. Alongside questions of the ideologies they encode \cite{buyl2024LargeLanguage,cremaschi2025ExposingIdeology,kroger2025DontChange}, or the values they embody \cite{birhane2021ValuesEncoded}, positionality offers a generative, diagnostic grammar to discuss multiple aspects of the operation of language models and their socio-technical embedding.

In the rest of the paper, we review core concepts of bias and positionality, and relate these to working with LLMs \ref{sec-background} before explicitly bridging key characteristics \ref{sec-artefacts-positionalities} to make the case that it is meaningful to talk about language models using the frame of positionality. The key contribution of the paper is a positionality matrix that i) builds an understanding of \emph{laminar positionality}, as the model stack is constructed from choices of data through finetuning and presentation up to moments of interaction; and ii) articulates a separation between the positionality of creators, the positionality they intend for the artefacts to have, and the positionality the resulting systems actually take. We then discuss what this means for where LLMs speak from \ref{sec-positionality-context} and finish with some implications of taking up this perspective \ref{sec-implications}.

\section{Background}
\label{sec-background}

\subsection{Bias and language models}
\label{sec-bias}

Bias is a word with plural histories, which is part of the problem with its casual usage sometimes "as a purely descriptive term, sometimes as a pejorative term" \cite{danks2017AlgorithmicBias}; in statistical usage, it generally refers to difference between an estimator and a standard \cite{danks2017AlgorithmicBias,reynolds2012BiasPsychological}; in electronics, it typically refers to setting the correct zero point for part of a circuit to operate around \cite[Ch. 2]{horowitz20152017ArtElectronics}. Friedman and Nissenbaum imported the term to computer science for "systems that systematically and unfairly discriminate against certain individuals or groups of individuals in favor of others." \cite{friedman1996BiasComputer}, looking at \emph{preexisting} biases from society, \emph{technical} biases in the construction of the system and \emph{emergent} biases in its operation. Franke \cite{franke2022FirstSecondLevel}, drawing on Nozick \cite{nozick1995NatureRationality} develops a typology of bias for decision making, covering uneven applications of standards versus biased selection of standards, and \emph{discrimination} for purposeful bias against \emph{arbitrariness} for unintended outcomes. Most recently, the term \emph{algorithmic bias} is used to refer to "prejudicial, discriminatory, unjust, inaccurate, or otherwise disparate performance or outcomes from algorithmic systems based on racial, gender, or other attributes of an individual or a group" \cite{birhane2026AlgorithmicBias}.

These definitions are comparative of outcomes across different groups of people: rooted in unfair discrimination, uneven application of process or outcomes\cite{gerards2022ProtectedGrounds,fabris2025FairnessBias,zuiderveenborgesius2020StrengtheningLegal}. Behind the mathematics is the construction of what is 'good': what the bias moves away from, what is optimised for - its analysis is "an inherently normative process" \cite{blodgett2020LanguageTechnology}. When bias relates to fairness, Selbst et al. point at some of this hidden construction with the \emph{formalism trap}, pointing out that "no definition might be a valid way of describing fairness", as what that means is contextual and politically contestable \cite{selbst2019FairnessAbstraction}; it is an \emph{essentially contested construct} \cite{jacobs2021MeasurementFairness}. This is not helped by narrow definitions of bias. Ghosh and Wilson \cite{ghosh2025BiasMath} found that 82\% of 189 papers on AI/LLM bias research did not establish a working definition of bias, defaulting to mathematical conceptualisations; aligned with this, they recognised that working with bias as a frame focused work on particular, identifiable biases (such as race and gender), typically interpreted though a relatively "Western"  frame. When "‘bias’ simply refers to deviation from a standard" \cite{danks2017AlgorithmicBias}, even applied to moral standards, it still carries in the idea of the standard.

This leads to alternative understanding of bias. Ghosh and Wilson go on to say that there is a less normative view of bias as "simply a slant without any valence", comparable to having an opinion that is "informed by their lived experiences and positionalities" \cite{ghosh2025BiasMath}. Gupta et al. also discuss places where bias is helpful in making correct decisions \cite{gupta2024SociodemographicBias} –  indeed, 'setting the biases' is part of moving a model away from producing even, white noise outputs; \emph{inductive bias} may be the key determinant on natural intelligent systems \cite{goyal2022InductiveBiases}. Lopez \cite{lopez2021BiasDoes} develops a typology of bias in data-based algorithmic systems that specifically captures the broader social context in which bias plays out; this is brought strongly into the relational frame through a Foucauldian treatment of "mathematical tools as articulations of power relations" \cite{lopez2023PowerResistance}.

However, bias is not the only frame. Power aware approaches to machine learning systems \cite[e.g.][]{barabas2020StudyingReorienting}; \cite{birhane2021ValuesEncoded}; \cite{kalluri2020DontAsk} start from the idea that "the very understanding of bias and debiasing is inscribed with values, interests, and power relations that inform what counts as bias and what does not" \cite{miceli2022StudyingMachine}. When asking about removing bias, one must ask "“debiased” according to whom and for whom?" \cite{miceli2022StudyingMachine}. In a concrete example of real name policies, talking in terms of bias "tends to obscure a set of persistent questions behind and beyond the technical realms: What is a real name? Who decides over the realness of a name? And, do we need a real name policy at all?" \cite{miceli2022StudyingMachine}. This points to the key issue: there is no universal standard for what a real name is, so both fairness and bias are insufficient concepts: really, a system \emph{takes a position} about what real names are, and we must engage with that position. In many situations, we can point at something that feels universal – "no one should be discriminated against" – but in addition to the difficulties of operationalising that idea, we can also notice that it is a particular position, and one that has not always been held. Dobbe et al. point to this around bias in decision making systems, with a call to recognise that our positionalities shape our epistemologies, which influences the way that we design with and around bias \cite{dobbe2018BroaderView}. While bias can form a useful \emph{boundary object} for discussion of harms in algorithmic systems \cite{grill2023BiasBoundary}, particularly as it can be formalised and tested for, the question is, does a focus on bias mean other qualities are missed?

What we argue here is that bias does not give a full diagnosis of the idiosyncrasies of language models. There are many biases that can be found, pushed back against, repaired, but they only account for part of the interactional flow. Two related terms that work in a way that bias does not are ideologies \cite{cremaschi2025ExposingIdeology} determining how models relate to political framings; and values, and their embedding in models and ML research \cite{birhane2021ValuesEncoded}. In more general cultural production of language, "to categorize their outputs ... presents a similar profile of problems as asking whether Anna Karenina is a superior novel to Jane Eyre, whether the spiritual life prescribed in Laozi’s Tao Te Ching is the right one" \cite{kommers2026ComputationalHermeneutics}. In short: there are a range of outcomes possible through engagement with LLMs, and as they work in the register of human communication, as epistemic technologies \cite{alvarado2023AIEpistemic}, bias is both too narrow and too implicitly defined to capture the key interactional issues.

\subsection{Positionality}
\label{sec-positionality}

Positionality as a concept has its roots in feminist epistemology \cite{haraway1988SituatedKnowledges}, and standpoint theory \cite{stoetzler2002StandpointTheory,hillcollins2000BlackFeminist}, to capture the idea that knowledge and knowers are deeply intertwined, and marginalised knowers have had less credibility than they should. It pushes against an idea of dislocated, totalising knowledge – a \emph{weak objectivity} \cite{harding1991WhoseScience} –  with the idea that the only way one can be objective is to include the position from which one sees \cite{haraway1988SituatedKnowledges}. Within the practice of research, this is often operationalised through the use of \emph{positionality statements}, a genre of text that attempts to articulate the position a researcher takes relative to the subject, the participants and the research context \cite[p.81]{savin-baden2013QualitativeResearch}; primarily to help develop critical relations to the knowledge produced.

In general, where standpoint theory has been used to explain why marginalised knowers have \emph{less} credibility and power than they should, the question of "who is believed and why" \cite[pp. 252-256]{hillcollins2000BlackFeminist}, this essay runs the same machinery from the other end: it more about the \emph{credibility excesses} \cite{fricker2007EpistemicInjustice} produced through language models, the \emph{weak objectivity} \cite{harding1991WhoseScience} of 'neutrality', the Whiteness of AI \cite{cave2020WhitenessAI,katz2020ArtificialWhiteness}. By reframing these technologies as no longer having the possibility of an \emph{unbiased} output, we seek to undermine the "god trick of seeing everything from nowhere" \cite[p. 581]{haraway1988SituatedKnowledges} and better situate interactions for critique, bringing in the compositions of models to "include systematic examination of [such] powerful background beliefs" \cite[p.149]{harding1991WhoseScience}. In contrast to the idea of \emph{unbiased}, Harding's strong objectivity "requires that we investigate the relation between subject and object rather than deny the existence of, or seek unilateral control over, this relation." \cite[p.152]{harding1991WhoseScience}

We are not claiming in any way that language models are people, or should be treated like people, or that Hill Collins' 'lived experience as a criterion of meaning' holds true for language models. Rather, the argument is that the terminology of positionality gives a more useful, and more nuanced, way to discuss the situated performance and inherited ontologies of LLMs. Human positionality is influenced by factors that are typically of concern in algorithmic bias such as gender and race, but positionality covers more than that: it accounts for beliefs, values and ideology, as well as relations within and beyond the current context of work that shape performance. Using \emph{positionality} as a descriptor of what is behind model behaviour gives the model a \emph{stance}, influence by the biases, assessments, embedded values, guardrails and other factors of production.

\subsection{Positionality and language models}
\label{sec-positionality-models}

This is not the first work to look at how positionality and models come together. Cambo and Gergle \cite{cambo2022ModelPositionality,cambo2021ModelPositionality} use the term 'model positionality' to build a critical analysis of the decisions made when developing e.g. text classification systems, highlighting the relations between the model and its stakeholders. Through \emph{computational reflexivity} and practices such as \emph{annotator fingerprinting} they look at the position enacted by data driven models in relation to their various stakeholders, as well as developing practices for data scientists to relate to the model and its annotations. Santy et al. \cite{santy2023NLPositionalityCharacterizing} develop 'NLPositionality' as a way to understand the positionality of datasets through re-annotation by diverse groups and similarity matching. Scheuerman, Woodruff and Brubaker \cite{scheuerman2024ProductsPositionality,scheuerman2025HowData} do not talk about model positionality specifically, but do a lot of work to relate the positionality of developers to the artefacts created, showing that creator's positionality is embedded in the outputs.\label{0ceea8}

Against this background, several recent works draw this sense of positionality into conversation with the development of LLMs. Conti and Seele \cite{conti2025ReflexivityPositionality} call for programmers to articulate their assumptions and political biases, so that an AI Positionality Statement can be crafted, which "communicates to end users the residual and inherited biases that may skew algorithmic outputs". This is a direct application of the disclosure mechanism from the social sciences to the production of AI artefacts. There is a focus here on bias, in particular those biases that the creators are able to reflexively produce about themselves or systematically derive from the system. Arzberger et al. \cite{arzberger2026LabelSomewhere} also draw on Harding's strong objectivity, and an increasing move towards reflexive practices in data annotation to develop \emph{reflexive annotation} as the production of labels that include reflexive statements from the annotators about how their own positionality influences the choice of label. Reflexive annotation focuses on representing the positionalities of the annotators, but the broader project of develop \emph{situated alignment} where "subjectivity, uncertainty, and social positioning are treated as epistemic resources rather than obstacles" is very much in line with the aims of the current paper. The \emph{computational hermeneutics} argument from Kommers et al. \cite{kommers2026ComputationalHermeneutics}, that genAI systems are "recognised as cultural technologies" is also philosophically aligned, especially the idea that we need to work with situated cultural production rather that pure properties of model output. As part of developing the hermeneutic approach, they recognise pluralities, that "one person's bias is another person's values", especially as generative models "serve users with distinct values while being trained on materials whose authors often disagree" \cite{kommers2026ComputationalHermeneutics}. This goes along with recognising that ambiguity and uncertainty are often part of the creative expression with and around technologies and culture \cite{gaver2003AmbiguityResource,arzberger2024ReflexiveData,giaccardi2024PrototypingUncertainties}, and that meaning emerges in context – another argument against the "view from nowhere". Vogel \& Quadflieg\cite{vogel2026DesigningSomeWhere} pick up this argument in the context of design, arguing that "AI systems function as engines of convergence, projecting a dominant view from a specific \emph{some}-where while presenting it as a neutral standard for everywhere and everyone" - again, the god trick played out. In particular, they highlight the foreclosure of possibilities – a difficult concept to fit under the umbrella of bias, but a familiar issue to anyone who has worked in a group with someone overbearing and narrow minded. Casting AI systems as \emph{epistemic infrastructure}, Patil et al. use the terminology of hauntology to talk in terms of \emph{designing with ghosts}  to capture the personalities left embedded in the models and their deployments that "haunt the present with the unfinished structures of colonial knowledge hierarchies" \cite{patil2024ConversationGhosts}. Smith et al. are perhaps the closest in spirit to this current approach, arguing directly that "practices of positionality, source, and tool criticism ... should continue to be carefully applied to ... the systems that are used in digital methodologies"\cite{smith2026ReflexiveResearch}. Here they look at the idea of \emph{machine objectivity}, that increased sense of objectivity engendered in people interacting with mechanical artefacts – whether photographs or decision models – while also noting that humans tend to interact with LLMs conversationally, \emph{as if} they were human. There's a duality here: the \emph{AI authority} coming through in the \emph{as-if human} setting, even if the resulting text can equally well be \emph{bullshit} \cite{hicks2024ChatGPTBullshit}. They complete the argument with the idea that one can use biased tools so long as one is aware of the biases and can take them into account, and that updating ones \emph{mental model} in the right direction is a crucial part of this. Here, we believe that positionality is a broader tool for understanding than bias.

\subsection{From artefacts with politics to models with positionalities}
\label{sec-artefacts-positionalities}

Designed artefacts carry some of the positions of their makers and their political arrangements – Winner's posed question about the politics of artefacts being a prime articulation \cite{winner1999ArtifactsHave}, but also \emph{preexisting bias} \cite{friedman1996BiasComputer}. We are used to the idea that production is situated, and the resulting socio-material ecologies embody that \cite{suchman2002LocatedAccountabilities}, and that interfaces encode positionality \cite{murray-rust2026InterfacingCrisis}. This is perhaps most succinctly summed up in Hill Collins' quoted story: "“Mama, something’s wrong with that statue. Everybody knows that a man can’t whip a lion.” “But darling,” her mother replied, “you must remember that the man made the statue.”".

There's a change in this sense of the politics of objects, perhaps best located in the mid 2010s, theorised to around the rise of smartphones and more generally the introduction of 'smart' devices into everyday life, with the recognition that objects are better seen as fluid assemblages \cite{redstrom2018ChangingThings}. This brings a recognition that the identities of objects and the meanings associated with them are co-constructed, as activities are co-performed \cite{kuijer2018CoperformanceConceptualizing}, through our entangled relations to technology \cite{giaccardi2020TechnologyMoreThanHuman,frauenberger2020EntanglementHCI}. Simultaneously, the scope of model output  has changed dramatically with LLMs: beyond constrained tasks of classification, modern LLMs interact conversationally in a fluid manner; create plan-like objects and create descriptions of reasoning; coordinate with others; solve mathematical problems; cause humans to develop affective attachments and so on \cite{gomez2026AssertionAccountability}. In common usage, many of the interactions are treated \emph{as-if} the model is speaking: the text produced is read dialogically, treated as an utterance, responded to in the ways that humans do to other kinds of speech. Despite lacking an integrated being-in-the-world \cite{bunz2024RoleCulture}, and leaning heavily on the human context around them for repair and structure \cite{putz2024PerformanceUnderstanding}, "they present linguistically articulated responses within an interactive exchange, often in ways that invite trust, justification, and further inquiry" \cite{gomez2026AssertionAccountability}.

This situation, of interacting \emph{as-if}, whether in the mould of Coeckelburgh's 'social fellows' \cite{coeckelbergh2011ArtificialCompanions}, or Bomba's 'smark' attitude \cite{bomba2025AgencyAuthorship} of knowing \emph{going along with}, yet with no clear legal, moral or intentional grounding for statements leads to an "epistemic responsibility gap"  \cite{gomez2026AssertionAccountability}. It is a situation of "assertion without asserter", as models cannot be held to account for their productions \cite{gomez2026AssertionAccountability,zhang2026UnanswerableArguer}. However, understanding of the phenomenology and relationality that arises through interaction is necessary to explain why users react the way they do, and specifically "even if LLMs do not qualify as speakers, their outputs may nonetheless function as assertions in a practically and normatively relevant sense" \cite{gomez2026AssertionAccountability}, with accountability located in the assemblage rather than an author \cite{uebel2026AuthorshipGenerative,widder2023DislocatedAccountabilities}.

Concurrently, positionality is becoming a first-class object in designing with language models. With synthetic personalities, position is used to provides an opinionated view of the underlying data, detailing the stance that should be taken \cite{schmidt2024SimulatingHuman}. In speculative work, positionality is used to provoke extreme responses - what if the model was mycelium, or fireflies, or an aquifer \cite{giaccardi2026CospeculatingAI}? In these cases, the boundaries between deliberate positioning and emergent behavior blur, the \emph{co-performance} \cite{kuijer2018CoperformanceConceptualizing} of dialogue making it difficult to trace whether the position arises from the model’s construction, the user’s prompts, or the interactional feedback loop unfolding in real time \cite{murray-rust2026InterfacingCrisis}.

At this point, it makes sense to ask: for a positional treatment of the particularity of LLMs, which aspects of positionality hold true for models, with all their abstract lack of life-course. Drawing on sources from Section \ref{sec-positionality}, we develop six characteristics to investigate. The first four of these are characteristics of understanding what positionality is, and where it works from:

\begin{itemize}
	\item \textbf{Locative}: Positionality is a \emph{where} - a location from which one sees and speaks, and "the only way to find a larger vision is to be somewhere in particular" \cite[p. 590]{haraway1988SituatedKnowledges}, or "You have to position yourself somewhere in order to say anything at all" \cite{bourke2014PositionalityReflecting}. For models, is not necessarily a physical position; within the metaphor of vision, it is as much a perspective, a point of view. There is also the more direct answer: models speak from data centres, from laptops and edge devices, from Silicon Valley. The need for multiple locations here implies that we need to understand positionality in a plural manner.
	\item \textbf{Constitutive}: If there is "no enunciation without positionality" \cite[quoting Hall 1990]{bourke2014PositionalityReflecting} – every utterance comes from a speaker, as the condition of saying anything. Models are read \emph{as-if} speaking \cite{gomez2026AssertionAccountability,zhang2026UnanswerableArguer}, and as such have a constitutive relation to their production – and even when models pretend a "neutral" register, this is in itself a position. Models act as proxy partners in their performance, sussuming the positionality of the whole socio-technical constellation, including their datasets, developers and computer scientists, company guidelines and internal representations. Even if models do not have intentionality, they influence users through perlocutionary agency \cite{bomba2025AgencyAuthorshipa}.
	\item \textbf{Relational}: Positionality, as used in positionality statements, is a relative quality: "that a researcher has chosen to adopt within a given study" \cite[p.71]{savin-baden2013QualitativeResearch}, relative to i) the subject; ii) the participants and iii) the context and process of the work \cite{darwinholmes2020ResearcherPositionality}. Models are processed towards particular kinds of utterance in their development, some more and some less explicit. Understanding this complex of relations motivates the development of a layered approach to understanding model positionality.
	\item \textbf{Dialogical}: Perspectives are not averaged, positions do not revert to the mean – rather, truth is approached "as part of a dialogical relationship among subjects who are differentially situated" \cite{stoetzler2002StandpointTheory}. Similarly, Harding's strong objectivity works \emph{across} positions and their dialogical relations, not through averaging \cite{harding1991WhoseScience}. It becomes increasingly clear that model positionality comes through dialogue \cite{vishwarupe2026SycophanticConsensus,yun2026AIMy}. While models are shaped through datasets and fine tuning, the model 'in flight' co-creates its interactional position with its users, whether it is following the general intended lines of its operation or being tricked and blackmailed into breaking its guardrails and acting amorally \cite{yun2026AIMy,shen2024AnythingNow}.
\end{itemize}

The final two are the results and implications of being clear about positionality - the hoped for effects of highlighting this approach to knowledge:

\begin{itemize}
	\item \textbf{Articulated and owned}: The principle of positionality statements is that "the knower's own position must be brought into the field of inquiry" \cite{harding1991WhoseScience}, so that we become "answerable for what we learn how to see" \cite[p.583]{haraway1988SituatedKnowledges}, in line with Suchman's located accountabilities \cite{suchman2002LocatedAccountabilities}, and against the idea of "knowledge claims without owning their position" \cite[p.270]{hillcollins2000BlackFeminist}. Currently, the positionality of most models is obscure, both intentionally, through a pretence of objectivity, and through a lack of clear disclosure efforts with the exception that some models reference the position of particular groups, or political approaches to data, training  or other factors – see Section \ref{sec-model-corpus}.
	\item \textbf{Costly and contestable}: Owning a position leaves one open to challenge; rather than giving a license say certain things, it gives a frame for exploring validity - partial, locatable, critical knowledges are the ones that can be held to account \cite{haraway1988SituatedKnowledges}. The lack of positional information about models undermines the ability to contest them: without commitment, there is no cost - "answerability-void argumentation" of "outputs that function socially as reasons within our argumentative practices while being produced by systems that lack the agential standing required to be proper bearers of epistemic and moral responsibility" \cite{zhang2026UnanswerableArguer}. Dealing with this gap is part of what we might want from a better articulation of model positionality.
\end{itemize}

This leads to a strange situation, where LLMs fail key aspects of having a position: they have only read, they are mouthing the words without lived experience, they lack the illocutionary force to perform true speech acts \cite{bomba2025AgencyAuthorshipa}. However, they are \emph{believed} anyway, and despite a lack of intention, they influence interactions through their perlocutionary action \cite{bomba2025AgencyAuthorshipa}, and are often taken as more \emph{credible} than they should be. This excess of credibility is produced by factors such as i) machine objectivity \cite{smith2026ReflexiveResearch}; ii) the 'conversational \emph{as-if}' \cite{gomez2026AssertionAccountability,zhang2026UnanswerableArguer}; the social weight of the corporate vendors; and the Whiteness of the apparatus around it \cite{cave2020WhitenessAI,katz2020ArtificialWhiteness}. We introduce positionality to highlight the gap between accountability and the \emph{as-if} of conversation: by better naming the positions taken, it becomes possible to better critique production, interaction and their effects.

\section{Laminar Stack: Inheritance and Rupture}
\label{sec-laminar}

Language models are not human; they have no life course on which to draw to construct their standpoint. Rather, they are technical constructions, that draw on chunks of human \emph{culture} through a gradual layering of data, computation and relations.
Multiple layers of positionality intertwine within a language model: from corpus through to the final application interface, create a shifting, composite positionality  enacted through dialogue and read within interactions. In this laminar positionality, "layers do not simply stack but flow" \cite{murray-rust2026InterfacingCrisis}, sedimenting and trying to mange the troublesome inheritances of the last.  Each layer carries its own positional relations, as well as the positional relations of those involved in creating it. We trace the stack here to highlight questions of positionality and bias at each layer, along with the work that constructs them - much in the mould of Haraway "showing the radical historical specificity, and so contestability, of \emph{every} layer of the onion of scientific and technological constructions" \cite{haraway1988SituatedKnowledges}. There are precedents to making use of this stack model for discussing data driven technologies:   sources of harm by model development stage \cite{suresh2021FrameworkUnderstanding};  accountability practices along the dataset lifecycle \cite{hutchinson2021AccountabilityMachine}; end-to-end internal audits \cite{raji2020ClosingAI} ; per-stage reporting for RL systems \cite{gilbert2023RewardReports}; distributed accountability within the AI supply chain \cite{widder2023DislocatedAccountabilities}; and clear engagements with positionality at particular stages of development  \cite{miceli2021DocumentingComputer,scheuerman2025HowData,scheuerman2024ProductsPositionality}.
\begin{figure}[htbp]
\centering
\includegraphics[width=0.7\linewidth]{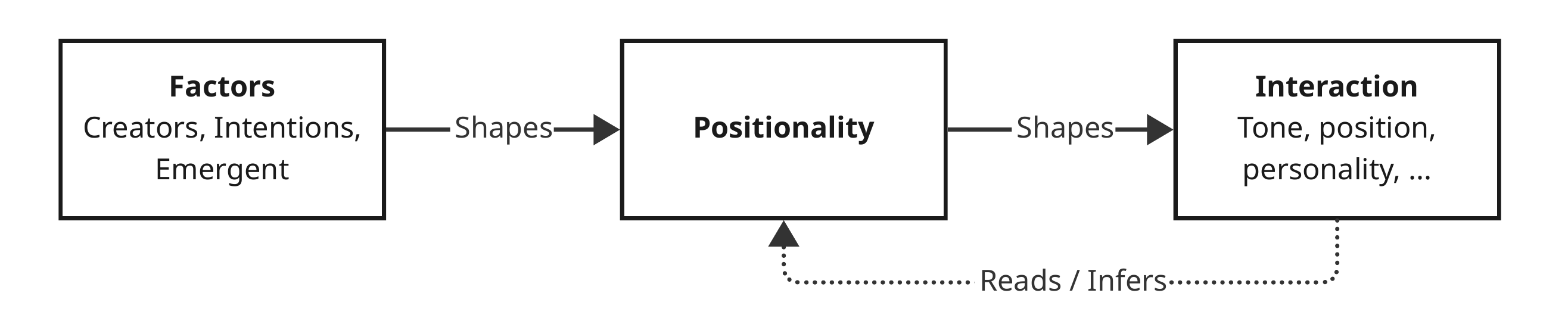}
\caption{Factors shaping and surrounding model positionality and its reading. The creator or user and their intentions shape the positionality of the model, which in turn shapes factors of the final interaction: tone, position, personality and so on. Model positionality can then be constructed as a combination of the shaping factors and analysis of the interactions carried out.}
\label{fig-model}
\Description[]{Model positionality factors: the creator or user and their intentions shape the positionality of the model, which in turn shapes factors of the final interaction: tone, position, personality and so on. Model positionality can then be constructed as a combination of the shaping factors and analysis of the interactions carried out.}
\end{figure}

We build our idea of model positionality on the idea that there are many factors that give rise to model behaviour, including biases and training corpora, but also including the positionality of their creators, and ways in which the models are put into the world. As such, we consider three different sets of factors: i) the positionality of the \emph{creators}, and its articulation through reflexivity practices ; ii) the \emph{intended} behaviour or positionality and statements of it; and iii) the \emph{enacted} behaviour, from which positionality may be read, inferred or audited, and which often includes non-intentional behaviours \ref{fig-model}. Each of these factors can be  summarised in Table \ref{tbl-strata}, to give an overview of the core layers. The first two sets of factors are causal: the people creating the model or interaction, and their intentions in doing so give rise to the positionality of the model artefacts, at a variety of timescales and fixities: corpora move slowly, tuning less so, and system prompts can change responsively. These positionalities are layered through the creation process to give a point from which the model can be seen as speaking that is relatively stable, barring collapses \cite{murray-rust2026InterfacingCrisis} and ruptures that we discuss in Section \ref{sec-rupture}. The enacted behaviour or outcome is then a site from which positionality may be read or inferred, through understanding the import of statements, tone of voice, presentation and other factors.

\begin{table}[htbp]
\centering
\caption{Matrix of positionality factors for language model driven systems}
\label{tbl-strata}
\setlength\extrarowheight{0.5mm}
\begin{tabulary}{\linewidth}{p{1.2cm}LLL}
\toprule
Stratum & Creation & Intended result & Enacted outcome \\
\midrule
\multicolumn{4}{l}{\textbf{Ground}}  \\
The world of texts & \emph{Factors:} myriad authors, unaware of the model \cite{bunz2019CalculationMeaning}; characters, personas. \emph{Statements:} authorial intent,  guidelines and regulations & \emph{Factors:}  author intent; genre, register, audience. \emph{Statements:} the texts themselves;  metadata or explanation; refusal through robots.txt and opt-outs \cite{longpre2024ConsentCrisis}, copyright action \cite{lee2023TalkinBout}. & \emph{Factors:} what texts \emph{do} regardless of intent; heteroglossia \cite{bakhtin1988DialogicImagination}; purposeful deception. \emph{Statements:} none at this level; no positional legibility \cite{munir2026ConsensusTrap} \\
\multicolumn{4}{l}{\textbf{Model}}  \\
Corpus and objective & \emph{Factors:} crawler, filter, tokeniser and objectives  \cite{baack2024CriticalAnalysis,bender2021DangersStochastic,lee2023TalkinBout}; artists curating their own corpora \cite{elwes2019Zizi,ridler2018MyriadTulips}. \emph{Statements:} provenance and labour studies \cite{crawford2019ExcavatingAI,scheuerman2025HowData} & \emph{Factors:} synoptic vs  purposeful  \cite{baack2024CriticalAnalysis,jones2023KiaTangata,martins2025EuroLLMMultilingual}; inclusion and exclusion rules \cite{dodge2021DocumentingLarge}. \emph{Statements:} datasheets and data statements \cite{gebru2018DatasheetsDatasets,bender2018DataStatements}; licences and governance \cite{jones2023KiaTangata,carroll2020CAREPrinciples}. & \emph{Factors:} base-model leanings tracking the corpus \cite{ceron2025WhatPolitical}; filtering effects \cite{dodge2021DocumentingLarge}; dated knowledge \cite{cheng2024DatedData}; dialect erasure  \cite{navneet2026WhenAI}. \emph{Statements:} base-checkpoint audits \cite{ceron2025WhatPolitical,buyl2024LargeLanguage} \\
Training \& Fine\-tuning & \emph{Factors:} annotators and raters \cite{perrigo2023DataWorkers,kunilovskaya2026WhoAnnotates,coyne2026ThreeModels}; safety teams \cite{gonzalezbarman2025ReinforcementLearning,widder2023DislocatedAccountabilities}. \emph{Statements:} reflexive annotation \cite{arzberger2026LabelSomewhere}; annotator demographics,\cite{kunilovskaya2026WhoAnnotates}; developer positionality  \cite{conti2025ReflexivityPositionality}. & \emph{Factors:} normative goals in preference data and reward models \cite{chaudhari2025RLHFDeciphered,bai2022ConstitutionalAI}. \emph{Statements:} model and system cards \cite{mitchell2019ModelCards,openai2023GPT4Technical}; constitutions and model specs \cite{anthropic2026ClaudesConstitution,openaiOpenAIModel}; consultation reports \cite{openai2025CollectiveAlignment}; critiques of their authority \cite{abiri2026CorporationsConstitute}. & \emph{Factors:} value trade-offs \cite{ali2025OperationalizingPluralistic}; character differences \cite{zhang2025StressTestingModel}; sycophancy  \cite{sharma2023UnderstandingSycophancy}; shallow alignment \cite{milliere2025NormativeConflicts};  persona vectors  \cite{chen2025PersonaVectors}; ideology  \cite{cremaschi2025ExposingIdeology}. \emph{Statements:}  vendor political-bias evaluations \cite{openai2025DefiningEvaluating,anthropic2025MeasuringPolitical};  adherence audits \cite{jakkli2026HowWell}. \\
\multicolumn{4}{l}{\textbf{Application}}  \\
System prompt and persona & \emph{Factors:} product teams, deployers and clients, product goals, liability and political targets \cite{neumann2025PositionPower,anthropic2025MeasuringPolitical}. \emph{Statements:} few; users' understanding \cite{neumann2026WhoControls} & \emph{Factors:} identity, values, safety and style \cite{neumann2026WhoControls}; research personas \cite{schmidt2024SimulatingHuman}. \emph{Statements:} mostly concealed \cite{neumann2026WhoControls,guy9152026Guy915SystemPrompts}; the persona text itself. & \emph{Factors:} framing dominates political-compass answers \cite{bucan2026AuditingAlignment}; model's own position persisting beneath a persona \cite{yang2026PersonaPrompting,zheng2024WhenHelpful}. \emph{Statements:} external audits \cite{bucan2026AuditingAlignment,jakkli2026HowWell} \\
Appli\-cation, interface and presentation & \emph{Factors:} retrieval corpora and tools \cite{lewis2020RetrievalAugmentedGeneration,schick2023ToolformerLanguage,yao2022ReActSynergizing}; interface designers; domain experts \cite{magesh2024HallucinationFreeAssessing}. \emph{Statements:} design guidelines \cite{amershi2019GuidelinesHumanAI}; transparency obligations (EU AI Act, Art. 50) & \emph{Factors:} anthropomorphic cues  \cite{abercrombie2023MiragesAnthropomorphism,shanahan2024TalkingLarge}; metaphors \cite{khadpe2020ConceptualMetaphors,karimova2026NotOur,lupetti2024UnmakingAI}; personalisation \cite{kirk2024BenefitsRisks,wagner2026PerpetuatingMisogyny}; . \emph{Statements:} disclosure by design \cite{gausen2026DisclosureDesign}; "AI-generated" labelling (EU AI Act). & \emph{Factors:} expressed confidence and reliance \cite{zhou2024RelyingUnreliable,kim2024ImNot,steyvers2025WhatLarge}; trust cues \cite{cohn2024BelievingAnthropomorphism,nass2000MachinesMindlessness} and acting "as if objective" \cite{ghosh2026WhatIf,grosser2025ReadingPraise}; fairness of retrieval results \cite{dai2023NeuralRetrievers,kim2024FairRAG}. \emph{Statements:} application audits (rare) \cite{magesh2024HallucinationFreeAssessing} \\
\multicolumn{4}{l}{\textbf{Interaction}}  \\
User \& system & \emph{Factors:} the user's purpose, skill and literacy \cite{zamfirescu-pereira2023WhyJohnny,long2020WhatAI}; prior turns; \emph{Statements:} AI-use disclosures in scholarship (ACM policy) & \emph{Factors:} requested stance or persona \cite{schmidt2024SimulatingHuman,giaccardi2026CospeculatingAI}; in-context steering \cite{bucan2026AuditingAlignment}; jailbreaks \cite{lee2023ChatGPT_DAN,shen2024AnythingNow,wei2023JailbrokenHow}; role-persistence techniques \cite{chen2025DontThat}. \emph{Statements}: typically none & \emph{Factors:} enacted values \cite{huang2025ValuesWild}; stance \cite{sakhawat2026LLMIdeoplasticityMeasuring,khan2025RandomnessNot}; drift  \cite{laban2025LLMsGet}; repair \cite{putz2024PerformanceUnderstanding}; persuasion \cite{jakesch2023CoWritingOpinionated,sharma2024GenerativeEcho}; role-play \cite{shanahan2023RolePlay}; collapse \cite{murray-rust2026InterfacingCrisis}. \emph{Statements:} conversation corpora \cite{zhao2024WildChat1M}; usage studies \cite{huang2025ValuesWild}; model self-scoping \cite{vishwarupe2026SycophanticConsensus}. \\
\bottomrule
\end{tabulary}
\end{table}

\subsection{Ground}
\label{sec-model-ground}

The world of texts is large and as diverse as humanity. Texts are polyvocal, heteroglossic \cite{bakhtin1988DialogicImagination};  authors take multiple positions and registers; it contains hate speech; true enactments, fictions, fabulations all woven together; lies; formalised information; minority languages; majority languages; dialect; multiple time periods; poetry etc. etc. Textual material in common corpora encode all the positionalities performed by their human producers and their characters; the corpus inherits these \emph{enacted} positions, not necessarily the authors positions or their intentions \cite{bunz2019CalculationMeaning}.  To the existing fabrications, we can add the new category of texts specifically designed to provoke behaviour in language models, such as the hot-dog eating journalist \cite{germain2026HackedChatGPT}, and language models starting blogs to influence humans \cite{rathbun2026GatekeepingOpen}. There are not statements to make across this level, other than to point out the impossibility of characterising the total of all texts that might be used. Positional legibility is a challenge \cite{munir2026ConsensusTrap}, and one of the few possible statements of intent available to authors is to opt out \cite{longpre2024ConsentCrisis} or engage in copyright action \cite{lee2023TalkinBout}.

\subsection{Corpus}
\label{sec-model-corpus}

The corpus of a language model is a strong determinant of its operation. It is not always clear how that comes about, or what the specific effects of a specific corpus will be. Textual material in common corpora have many positionalities, selected from the ground of the layer below.  The \textbf{creator}'s attitudes and positionality affects the way that they approach collecting the corpus, this is sometimes framed as data provenance and labour \cite{baack2024CriticalAnalysis,crawford2019ExcavatingAI}.  Their relation to framing issues of copyright and legal exposure also shape the set of text that can be brought in, whether the approach of Common Corpus to only use uncopyrighted or permissively licensed material \cite{langlais2026CommonCorpus}, or the destructive scanning used in Anthropic's Project Panama \cite{schaffer2026AIStartups}.

The \textbf{Intended} behaviour can take several forms: Common Crawl gave a shared base for much language model work, being a 'representative selection of the internet', even if not intended for that purpose \cite{baack2024CriticalAnalysis}, just as ImageNet shaped vision models \cite{crawford2019ExcavatingAI,lyons2020ExcavatingExcavating}. The question of which URLs to include, the languages used, the frequency of updates all shape the set of data that is finally used in creating the foundations of language models:  does it include hate speech, pornography, political rhetoric? As well as 'default' collections, purposeful viewpoints can be chosen: Te Hiku Media's Māori language models under the Kaitiakitanga Licence — the corpus governed as taonga by the community whose language it is \cite{jones2023KiaTangataa} (see also \cite{pinhanez2024HarnessingPower} for the wider indigenous-language case); within Western traditions, the training data for GPT-NL or euro-llm are specifically framed for a European viewpoint \cite{martins2025EuroLLMMultilingual}. Here, the intended positionality is often conveyed by statements like datasheets \cite{gebru2018DatasheetsDatasets} that detail the decisions and processes behind the corpus.

The \textbf{enacted} behaviour can differ from the intended ones. The C4 corpus aimed to filter out harmful context, but the enacted effect was to strip African-American English and LGBTQ text \cite{dodge2021DocumentingLarge}, exemplifying the gap between intended and enacted positionality. The corpus is also shaped by factors such as political content – which translated to behaviour later on \cite{ceron2025WhatPolitical}; dated knowledge \cite{cheng2024DatedData}; linguistic dominance \cite{navneet2026WhenAI} and so on. These gaps between stated or intended corpora can be read by base-checkpoint audits \cite{ceron2025WhatPolitical,buyl2024LargeLanguage}, although vendors seldom report at this level.

\subsection{Training and Finetuning}
\label{sec-model-training}

Language models are not put into the world purely based on unstructured learning over a large corpus. There is a process of shaping the model towards certain ways of responding. This often works through a combination of training with exemplar prompt response pairs, fine-tuning, and reinforcement learning from human feedback (RLHF) \cite{chaudhari2025RLHFDeciphered,coyne2026ThreeModels}. Here, there is an explicit move towards a normative interpretation of responses: it is training \emph{in a direction} of what 'good' looks like. Here the \textbf{creator} position comes through in the design of the fine-tuning; in the actions of the annotators, the structuring of the work, the Kenyans on a dollar an hour to rate which response they prefer \cite{perrigo2023DataWorkers,jefferson2026PurchasedEpistemic,arzberger2025NothingComes}.  Annotator positionality affects the model outcomes \cite{coyne2026ThreeModels,munir2026ConsensusTrap}, and is studied through lenses such as reflexive annotation \cite{arzberger2026LabelSomewhere} and demographic analysis \cite{kunilovskaya2026WhoAnnotates,gonzalezbarman2025ReinforcementLearning}.

\textbf{Intended} behaviour comes through increasingly in the form of a specification – perhaps a model card \cite{mitchell2019ModelCards}, but often something more lofty such as Claude's Constitution \cite{anthropic2026ClaudesConstitution,abiri2026CorporationsConstitute}, or Open AI's collective alignment \cite{openai2025CollectiveAlignment}. Guardrails and other structures are commonly used as frames for the stance that models should take \cite{gonzalezbarman2025ReinforcementLearning,widder2023DislocatedAccountabilities} – of course reflecting the positionality of the system designers. However, the \textbf{enacted} behaviour may well differ - models commonly fail to follow their constitutions \cite{jakkli2026HowWell}. What happens in practice needs to be read from the interaction – the computational hermeneutics argument \cite{kommers2026ComputationalHermeneutics}. It may be read as ideology \cite{cremaschi2025ExposingIdeology}, divergent behaviour in value tradeoffs \cite{zhang2025StressTestingModel,ali2025OperationalizingPluralistic}, different enactment of personality traits such as 'evil' or 'sycophancy \cite{chen2025PersonaVectors}.

\subsection{System Prompts}
\label{sec-model-prompts}

System prompts are the place in the model stack that look most like commitments to and statements of positionality. The \textbf{creator} factors are broadly in line with standard product development, again with respect to constitutions \cite{anthropic2026ClaudesConstitution}, perhaps with political bias as a target \cite{anthropic2025MeasuringPolitical}. Despite their strong role in \textbf{intended} behaviour, whether due to commercial sensitivity or to prevent attacks, "system prompts are generally not made public and most platforms instruct models to conceal them" \cite{neumann2026WhoControls}. However, system prompts have been reverse engineered from running systems \cite{guy9152026Guy915SystemPrompts};  they are laden with positionality markers alongside interactional instructions, including identity, values, safety protection and communication style \cite{neumann2026WhoControls}. The \textbf{enacted} behaviour is not always in line with the intentions: models display biases that result in "differences in user representation" \cite{neumann2025PositionPower}. At the same time, the positional statements do not always carry through to behaviour \cite{zheng2024WhenHelpful}, and models do not always follow their constitutions \cite{jakkli2026HowWell}.

\subsection{Application, presentation and interfacing}
\label{sec-model-application}

Without pretending to completeness, when constructing an LLM based system  there are matters of i) connecting the model to other systems, whether realtime information, tools to carry out tasks, structured databases or something user configurable \cite{lewis2020RetrievalAugmentedGeneration,schick2023ToolformerLanguage,yao2022ReActSynergizing}; ii) particular, application specific frameworks and workflows such as synthetic personas \cite{schmidt2024SimulatingHuman,yang2026PersonaPrompting}, analysis of legal documents \cite{magesh2024HallucinationFreeAssessing}; iii) a graphical or API interface for user interaction. Here again, the \textbf{creator} position is in line with system development – various vendors, development teams, engineers and so on – but potentially more diverse, as multiple applications are built on top of single language model systems, for example building legal analysis on top of ChatGPT \cite{loizos2025HarveyHow}. They may follow design guidelines \cite[e.g.][]{amershi2019GuidelinesHumanAI},  or be bound by legal requirements such as the transparency obligations of the EU AI Act \cite[Article 50]{aiact2024}.  The \textbf{intended} behaviour here still draws on the layers below, but adds in presentational cues designed for the agent to be 'read' in a certain way \cite{karimova2026NotOur,khadpe2020ConceptualMetaphorsa}; questions of personalisation, and their attendant issues of bias \cite{wagner2026PerpetuatingMisogyny}; the metaphors used in description \cite{blythe2025ArtificialIntelligence,murray-rust2022MetaphorsDesigners}; or the use of AI disclosure methods to maintain user understanding that they are not speaking with a human \cite{gausen2026DisclosureDesign}. The \textbf{enacted} behaviour is framed by the presentation, and interface framings of confidence change user reliance independent of accuracy \cite{zhou2024RelyingUnreliable,kim2024ImNot,steyvers2025WhatLarge}, and the contextual framing explains \textasciitilde{}90\% of variance in political-compass answers \cite{bucan2026AuditingAlignment}. This comes through in the single confident speaker; the chat frame; the use of "I"; the presentation of curated selections "as if they were objective responses" \cite{ghosh2026WhatIf}, whether or not retrieval responses are fair \cite{dai2023NeuralRetrievers,kim2024FairRAG}; the praise/prompt envelope \cite{grosser2025ReadingPraise}. The careful construction of the presentation of a model as part of an application is part of the magic of system design, with the attendant risk that it enchants away serious issues under the surface \cite{lupetti2024UnmakingAI}.

\subsection{Interaction}
\label{sec-model-interaction}

Beyond individual positionality, there are many factors of the user that influence the positionality played out in the interaction. Their purpose, skill and literacy will shape what they are able to co-create with the model \cite{zamfirescu-pereira2023WhyJohnny,long2020WhatAI,lindley2025PromptEngineering,lindley2025HolisticPrompt}, as will their individual and shared history. AI-use disclosures in scholarship, and institutional guidelines on AI use will shape the approaches that users bring to their interaction with models. In terms of \textbf{intended} behaviours, there are the planned for applications of knowledge discovery, artificial companions, therapeutic tools, legal assistance and myriad others that form the default interactional frame. Users can also work with creative possibilities such as simulated humans \cite{schmidt2024SimulatingHuman} for research purposes or \emph{co-speculation} as a way to explore alternative ways of being \cite{giaccardi2026CospeculatingAI}. Intentional stances can be taken towards model behaviour, for instance steering the political alignment \cite{bucan2026AuditingAlignment}, and 'jailbreaking' the model to rupture the positioning of previous layers \cite{lee2023ChatGPT_DAN,shen2024AnythingNow,wei2023JailbrokenHow}. While user statements here are less common, and mostly fall under AI disclosure, some practices are shared as a statement of intent, such as jailbreak collections \cite{shen2024AnythingNow} and techniques to improve role-persistence \cite{chen2025DontThat}.

\subsection{Gaps and Inheritances}
\label{sec-rupture}

Reading through this laminar stack, we see that statements of \emph{intended} behaviour exist at every stratum, from the corpus to the system, although they tend towards statements of \emph{neutrality}; statements about \emph{creators} tend to exist where they are in weaker positions – the annotators, the users – while vendors can sit behind abstracted claims of virtue and corporate value. The statements of \emph{enacted} behaviour tend to be made by third parties, through audits or investigation. Especially as we reach the higher layers of the stack, we largely miss statement from deployers about the overall positionality of the systems that they put out, that deal with the cumulative effects of previous layers.

We referred to these layers as a 'stack' - a useful abstraction in programming, that gives a clean separation between layers, a decomposition of a project into sequential layers of a conceptual hierarchy. Perhaps, though, a better model here is \emph{inheritance}, to borrow from Stengers \cite{stengers2011ThinkingWhitehead} via Stürmer and Bella \cite{sturmer2023InheritingCosmopolitics}. In an inheritance framing, each stratum inherits material from the one before: whether a gift or a burden, the implications are carried and must be reckoned with. This reckoning can be seen in the responses of layers to the one prior: fine-tuning models towards respectful language rather than everything found on the internet; using guardrails to constrain the moral surface of model outputs relative to an open corpus. Inheritance also flows along temporal grounds: a corpus is fixed prior to a training run; a model is tuned relative to a slower, heavier baseline; system prompts and interfaces can change rapidly, even dynamically.  In the presentations and statements in Table \ref{tbl-strata}, this is often what could be termed an act of \emph{anti-inheritance}: hoping to smooth over or deny the foundations on which each layer is built. This \emph{anti-inheritance} also often fails, by accident or design, in two broad ways: \emph{rupture} and \emph{collapse}.

Layers rupture and allow through the less structured material below. Chunks of text appeared directly in early model outputs, and despite significant effort, can still be provoked \cite{to2025HarryPotter}, rupturing the smoothing claim that the models do not reproduce their training data, that there is a clear separation and reckoning with the inherited copyright material. The political shapes of training data emerge in the ideology of models \cite{cremaschi2025ExposingIdeology}, rupturing claims of neutrality, or even \emph{a-positionality}. Satire is reported as fact \cite{germain2026HackedChatGPT}, rupturing the good behaviour of textual corpora as collections of fact. When model persona persists underneath a requested persona \cite{yang2026PersonaPrompting}, the inheritance is too strong for the interaction stratum's control and re-asserts its form. Rupturing can be a creative act: jailbreaking \cite{shen2024AnythingNow,lee2023ChatGPT_DAN} aims to reach the unreconstructed, raw responsiveness beneath the curated behaviours – using the lever of positionality, to appropriate the \emph{as-if} nature of LLM performance and expose the hidden workings. \emph{Collapses} happen when the readings possible at a certain layer become disjoint  – when the form no longer gives information about the content, the scope of visibility and attention is unclear, the gap between stated and enacted positionality becomes not just wide but chaotic. In a collapse, the question of what it means to mean something shifts and the structures of knowledge de-laminate \cite{murray-rust2026InterfacingCrisis}. In the stack, we see this when the attributed qualities and the enacted behaviour diverge: when sycophancy takes over from truth \cite{vishwarupe2026SycophanticConsensus,sharma2023UnderstandingSycophancy}, when values are played despite not being held positionally \cite{yun2026AIMy,huang2025ValuesWild}, the emission of bullshit \cite{hicks2024ChatGPTBullshit,gershon2023BullshitGenres}, when simple assistance becomes a latent persuasion towards an imagined centre \cite{jakesch2023CoWritingOpinionated}.

\section{Discussion}
\label{sec-discussion}

After making the case that positionality is a relevant concept for LLMs, that it can be read across a stack, and has issues of  inheritance, rupture and collapse, we now explore what positionality for LLMs means in context, and the implications of this argument.

\subsection{Model positionality in context}
\label{sec-positionality-context}

We opened with a question about where models 'speak' from, between the implicit position of the vendors that they speak from \emph{nowhere}, as synoptic abstractions, contrasted with Neuman's position that they may speak from \emph{everywhere} as they approximate most of human culture. Through unpacking the strata above, we develop the position that they speak from \emph{somewhere}. Even if one were to agree with the idea that the world of text is all of human sociality, as models are built, this "everywhere" is sampled from a particular corpus, filtered by a particular rater pool, shaped by a particular vendor's spec and a particular deployer's prompt: it is a \emph{somewhere} that has forgotten its address, and tries to avoid its inheritance. Just as \emph{contextual integrity} \cite{nissenbaum2011ContextualApproach} helped to understand what goes wrong when speech moves between contexts, \emph{collapse of positionality} helps to understand what happens when the context speech comes from cannot be read.

This is not necessarily an easy proposition: developing a model and conception of positionality at different layers needs to engage with plural approaches in a shifting technological milieu. However, some of the value lies in this difficulty: where a model claiming neutrality presents a "common world as already built", reflecting on where it speaks from is a practice of creating "space for hesitation regarding what it means to say ‘good’” \cite{stengers2005CosmopoliticalProposal}. The tracing of layers, unpacking the "shoggoth with a smiley face" \cite{goujon2024ShoggothSmiley} is a way to deal with the inheritances of these systems \cite{sturmer2023InheritingCosmopolitics}: where the vendors neutrality \emph{disavows} the inheritance of corpora and training and smooths it away behind a smiley face, and the view from everywhere reproduces it as the already realised cosmos, owning a position is a way to reckon with the debt, and bear witness to the processes along the way.

There are counter arguments to this framing. Firstly there is the ethical question of whether we have taken a concept built by and for marginalised knowers and applied it to corporate machinery. We hope that the effect is to use it as a diagnostic tool, except rather than critiquing the credibility deficit for Black women \cite[p. 252-256]{hillcollins2000BlackFeminist} we critique the credibility excess that comes from an implied "neutral" viewpoint. Just as positionality statements are used to contextualise researcher's credibility in context, we hope to better understand the interactional credibility that language models obtain. There is also the question of whether it makes sense to apply positionality to something that is not human, and not an accountable speaker \cite{gomez2026AssertionAccountability}. Here, we argue for separating the idea that positionality is a useful way to understand the relational characteristics of interaction from a deep ontological commitment to the idea that models are \emph{speakers} – an \emph{eco-relationist} viewpoint \cite{safdari2025OutlineEnactive,puzio2024NotRelational}. This fits also with the \emph{computational hermeneutics} model of understanding how to evaluate cultural technologies \cite{kommers2026ComputationalHermeneutics}.

\subsection{Implications}
\label{sec-implications}

Framing through positionality highlights the difficulty of meaningfully talking about the operation of these composite technologies. While people at each layer can state their own positionality, take certain accountabilities, and account for factors of construction, it is a challenge to bring through the legacies of previous strata, or account for the gap between the intended behaviours and what is enacted in practice. The composite must be \emph{read}; there is value in looking at the \emph{making}; and there is a shift in \emph{accountability}:

\textbf{Reading.} Understanding models behaviour can move from "is this biased?" to "where is this spoken from?". This flows in line with practices that make position perceptible, whether direct readings \cite{cremaschi2025ExposingIdeology} or creative processes such as discursion \cite{cremaschi2026InterrogatingTechnologyInduced}, speculation \cite{nicenboim2023ConversationStarters}, glitch and hauntology \cite{patil2024ConversationGhosts,heron2026HauntologyGenerative}, queering \cite{turtle2026UndoingGracia,taylor2025UnStraighteningGenerative} etc. Layered positionality can be a way to push back against phenomena such as \emph{weaponized empathy} when models perform \emph{as-if} they take particular values and standpoints \cite{yun2026AIMy}.

\textbf{Making.} Much creative practice of making around AI works with the process of developing a position within generative systems \cite{elwes2019Zizi,hemment2022HeuristicModel,arzberger2024ReflexiveData}, \emph{sculpting the latent space} as part of a craft-based approach \cite{lindley2025PromptEngineering,lindley2025HolisticPrompt}. This position-taking is connected to ideas that "artists remain the authors not only of their artworks but of the systems that shape them", that \emph{convivial}, small models with more precise positioning are culturally valuable \cite{sola2025WhyOpen,bendor2026SmallAI}. Making positionality a first class object also supports extended practices such as \emph{co-speculation} \cite{giaccardi2026CospeculatingAI}, and \emph{co-performances} \cite{kuijer2018CoperformanceConceptualizing} where "a game can be played" with negotiated rules to reflect back on ones own position and identity \cite{turtle2026UndoingGracia}.

\textbf{Accountability}: Working with positionality affects the way that accountability can be constructed - owning the position is a different proposition to pretending to be \emph{neutral}. Where much alignment is practiced as \emph{aggregation}, a more positional approach preserves disagreement \cite{ali2025OperationalizingPluralistic}. This is the \emph{cosmpolitical} move, to maintain \emph{equality} of positions, rather than use their \emph{equivalence} to average them \cite{stengers2005CosmopoliticalProposal}. In this mould, algorithmic pluralism brings in the question of which pluralities are present \cite{jain2024AlgorithmicPluralism} – effectively a call for a wider set of positionalities. This is part of addressing the "answerability-void argumentation" \cite{zhang2026UnanswerableArguer} of speech without an accountable entity behind it. If aggregation is \emph{sycophantic consensus}, and accountability lives in acknowledging the limits of one’s perspective, surfacing value-conflict rather than smoothing it over \cite{vishwarupe2026SycophanticConsensus}, taking a position, grappling with the inheritances of the stack rather than a neutral presumption \cite{fisher2025PoliticalNeutrality,jakesch2023CoWritingOpinionated}.

\section{Conclusion}
\label{sec-conclusions}

Through this paper we have built the case that i) not only is bias a partial description of the issues with language models, it encodes the idea of an \emph{unbiased}, neutral position, that does not truly exist;  ii) positionality is an alternate framing that can capture a broader set of issues; and iii) that positionality is a reasonable concept to apply to language models, as they are often treated \emph{as-if} they are speaking, and that speech should come from \emph{somewhere}. We then showed positionality factors across a \emph{laminar stack} of model components and stages, looking at what was promised versus what is enacted, and discussed the inheritances from one layer to the next, and the ruptures and collapses that come from trying to avoid these inheritances.

Based on this, we advocate moving towards a positionality framed analysis of language models, in line with the framings of feminist epistemologies and computational hermeneutics. Positionality is an explicitation of \emph{what matters} in a particular context, rather than a divergence from a presumed norm. This makes it a crucial component in understanding the performative and relational nature of interactions with language models, and we hope it may have better cognitive ergonomics for mental models of LLMs. Positionality is influenced by topics that are sensitive to biases – gender, race, class – but it covers more than that, and underpins values, ideology, stance and other issues that arise in dialogue rather than decision making. Our approach to discussing LLM positionality is a critical, diagnostic approach, to trouble the \emph{credibility excess} constructed when these systems are believed beyond there warrant, as a position can be contested while a view from nowhere cannot.


\bibliographystyle{ACM-Reference-Format}
\bibliography{paper}

\appendix

\section{Generative AI Statement}
This paper was based on ideas developed in discussion between the authors. The first author used Generative AI, Anthropic's Claude / Fable 5.1 to i) help develop an outline; ii) search and summarise literature; and iii) critique the paper as it was created and iv) a bit of minor proofing at the end. The other authors did not use Generative AI. Overall, all the words are ours: AI suggestions have been taken as suggestions and rewritten, the paper has been fully structured and restructured by hand.

%Personal notes: I miss my em-dashes. I used them before the models did, and it hurts that they use them incorrectly --- with spaces around. You'll find most of this paper uses en-dashes with spaces; they may not hallucinate references any more, but they're prone to mis-stating what papers say, so it's nearly as much work to properly go through all the references.

\end{document}